\documentclass{article}

\usepackage{microtype}
\usepackage[dvipdfmx]{graphicx}
\usepackage{subfigure}
\usepackage{booktabs} % for professional tables
 
\usepackage{hyperref}
\usepackage{amsmath}
\usepackage[accepted]{icml2018}

\icmltitlerunning{
Exploiting the Potential of Standard Convolutional Autoencoders %Standard ConvNet Architectures
for Image Restoration by Evolutionary Search
}

\begin{document}

\twocolumn[
\icmltitle{
Exploiting the Potential of Standard Convolutional Autoencoders\\
%Standard ConvNet Architectures\\
for Image Restoration by Evolutionary Search
%Have We Fully Explored the Architectural Space of Standard ConvNets?: Unveiling Their Potential for Image Restoration by Evolutionary Search
}
%Evolutionary Convolutional Autoencoders for Image Restoration}



\icmlsetsymbol{equal}{*}

\begin{icmlauthorlist}
\icmlauthor{Masanori Suganuma}{goo,to}
\icmlauthor{Mete Ozay}{goo}
\icmlauthor{Takayuki Okatani}{goo,to}
\end{icmlauthorlist}

\icmlaffiliation{goo}{Graduate School of Information Sciences, Tohoku University, Sendai, Japan}
\icmlaffiliation{to}{RIKEN, Tokyo, Japan}
%\icmlaffiliation{ed}{School of Computation, University of Edenborrow, Edenborrow, United Kingdom}

\icmlcorrespondingauthor{Masanori Suganuma}{suganuma@vision.is.tohoku.ac.jp}
\icmlcorrespondingauthor{Mete Ozay}{mozay@vision.is.tohoku.ac.jp}
\icmlcorrespondingauthor{Takayuki Okatani}{okatani@vision.is.tohoku.ac.jp}

%\icmlkeywords{Machine Learning, ICML}

\vskip 0.3in
]



\printAffiliationsAndNotice{}  % leave blank if no need to mention equal contribution

\begin{abstract}
Researchers have applied deep neural networks to image restoration tasks, in which they proposed various network architectures, loss functions, and training methods. In particular, adversarial training, which is employed in
recent studies, seems to be a key ingredient to success.
In this paper, we show that simple convolutional autoencoders (CAEs) built upon only standard network components, i.e., convolutional layers and skip connections, can outperform the state-of-the-art methods which employ adversarial training and sophisticated loss functions.
The secret is to employ an evolutionary algorithm to automatically search for good architectures.
Training optimized CAEs by minimizing the $\ell_2$ loss between reconstructed images and their ground truths using the ADAM optimizer is all we need.
Our experimental results show that this approach achieves $27.8$ dB peak signal to noise ratio (PSNR) on the
CelebA dataset and $40.4$ dB on the SVHN dataset, compared to $22.8$ dB and $33.0$ dB provided by the former state-of-the-art methods, respectively.
\end{abstract}

% section 1
\section{Introduction}
The task of image restoration, which is to recover a clean image from
its corrupted version, is usually an ill-posed inverse problem. In order to
resolve or mitigate its ill-posedness, researchers have incorporated
image priors such as edge statistics \cite{fattal}, total variation
\cite{perrone_total_2014}, and sparse representation
\cite{ksvd,yang_image_2010}, which are built on
intuition or statistics of natural images. Recently, learning-based methods which use
convolutional neural networks (CNNs) \cite{lecun_gradient_1998,krizhevsky_imagenet_2012} were introduced to overcome the limitation of these
hand-designed or simple priors, and have significantly improved
the state-of-the-art. 

In these studies, researchers have approached the problem mainly from
two directions. One is to design new network architectures and/or new loss functions. The other is to develop new training methods, such as the employment of adversarial training \cite{gan}.
Later studies naturally proposed more complicated architectures to improve the performance of earlier architectures.
Mao et al. \yrcite{red} proposed an architecture consisting of a chain of symmetric
convolutional and deconvolutional layers, between which they added skip
connections \cite{srivastava_2015,res}.
Tai et al. \yrcite{mem} proposed an $80$-layer memory network which contains a recursive unit and a gate unit.
Yang et al. \yrcite{yang2017high} proposed an image inpainting framework that uses two networks: one for capturing the global structure of an image, and one for reducing the discrepancy of texture appearance inside and outside missing image regions.
While many studies employ the $\ell_2$ distance between the clean and recovered images, 
some propose to use new loss functions such as the perceptual loss to obtain perceptually better results \cite{Johnson2016Perceptual,ledig2016photo}.

There are also studies on the development of new training methods.
A recent trend is to use adversarial training, where two networks are trained in an adversarial setting; a generator network is trained to perform image restoration, and a discriminator network is trained to distinguish whether an input is a true image or a restored one. 
The first work employing this framework for image inpainting is the {\em context encoder} of Pathak et al. \yrcite{pathak2016context}. 
They minimize the sum of a reconstruction loss over an encoder-decoder network for restoring intensities of missing pixels and an adversarial loss over additionally a discriminator network. 
Iizuka et al. \yrcite{IizukaSIGGRAPH2017} proposed an improved framework in which global and local context discriminators are used to generate realistic images.
While the above studies require the shapes of missing regions (i.e., masks) for training, Yeh et al. \yrcite{semantic} proposed a method which does not need masks for training.
Their method first learns a latent manifold of clean images by GANs and search for the closest encoding of a corrupted image to infer missing regions.
Despite its success in various application domains, GANs have several issues, such as difficulty of training
(e.g., mode collapse), difficulty with evaluation of
generated samples \cite{arXiv1711.10337}, and theoretical limitations \cite{Arora_ICML17_Generalization}.

A question arises from these recent developments: {\em what is (the most) important of these ingredients, i.e., the design of network architectures, loss functions, and adversarial training?} In this study, we report that convolutional autoencoders (CAEs) built only on standard components can outperform the existing methods on standard benchmark tests of image restoration.
We do not need adversarial training or any sophisticated loss; minimization of the standard $\ell_2$ loss using the ADAM optimizer \cite{adam} is all we need.
We instead employ an evolutionary algorithm \cite{suganuma} to exploit the potential of standard CAEs, which optimizes the number and size of filters and connections of each layer along with the total number of layers. 
The contribution of this study is summarized as follows:
\begin{itemize}
\item We show that simple CAEs built upon standard components such as convolutional layers and skip connections can achieve the state-of-the-art performance in image restoration tasks. Their training is performed by minimization of a standard $\ell_2$ loss; no adversarial training or novel hand-designed loss is used.
\item We propose to use an evolutionary algorithm to search for \textit{good} architectures of the CAEs, where 
the hyperparameters of each layer and connections of the layers are optimized.
\item To the best of our knowledge, this is the first study of automatic architecture search for image restoration tasks. Previous studies proposed methods for image classification and tested them on the task.
\end{itemize}

% Section 2
\section{Related Work}

\subsection{Deep Learning for Image Restoration}

Deep networks have shown good performance on various image restoration tasks, such as image denoising \cite{xie_image_2012,zhang_beyond_2017}, single image super-resolution (SISR) \cite{dong2014learning,ledig2016photo}, deblurring and compressive sensing \cite{xu2014deep, kulkarni_reconnet_2016,mousavi_learning_2017}, in addition to those mentioned in Sec.1. 
%All these studies propose new network architectures and/or new training methods. 
In particular, recent studies tend to rely on the framework of GANs \cite{gan} for training to improve accuracy or perceptual quality of restored images, e.g., \cite{pathak2016context, semantic}.

\subsection{Automatic Design of Network Architectures}

Neural networks have been and are being designed manually, sometimes with a lot of trial and error. Recently, an increasing attention is being paid to  automatic design of network architectures and hyperparameters \cite{miikkulainen_evolving_2017,xie_genetic_2017,liu2017progressive,smash,hierarchical}. The recent studies are roughly divided into two categories; those based on evolutionary algorithms and on reinforcement learning.

The employment of evolutionary algorithms for neural architecture search  has a long history \cite{schaffer_combinations_1992,stanley_evolving_2002}. In the past, the weights and connections of neural networks are attempted to be jointly optimized, whereas in recent studies, only architectures are optimized by evolutionary algorithms, and their weights are left to optimization by SGD and its variants.
Real et al. \yrcite{real_large_2017} showed that evolutionary algorithms can explore the space of large-scale neural networks, and achieve competitive performance in standard object classification datasets, although their method relies on large computational resources (e.g., a few hundred GPUs and ten days). Suganuma et al. \yrcite{suganuma} proposed a designing method based on cartesian genetic programming \cite{miller_cartesian_2000}, showing that architectural search can be performed using
two GPUs in ten days.

Another approach to neural architecture search is to use reinforcement learning. There are studies that employ the REINFORCE algorithm, policy gradient, and Q-learning to learn network topology \cite{zoph_neural_2016,baker_designing_2016,zhong_practical_2017,zoph2017learning}. These reinforcement learning-based approaches tend to be computational resource hungry, e.g., requiring 10-800 GPUs. 

In this study, we employ the method of Suganuma et al. \yrcite{suganuma} due to its computational efficiency, although we think that other recent light-weight methods could also be employed.
As their method was tested only on classification tasks as in other similar studies, we tailor it to designing CAEs for image restoration tasks. As will be described, we confine the search space to symmetric CAEs, by which we make it possible to design competitive architectures with a limited amount of computational resource (using 1 to 4 GPUs in a few days).

\subsection{Evaluation Methods for Image Restoration}

There is a growing tendency that perceptual quality rather than signal accuracy is employed for evaluation of image restoration methods \cite{ledig2016photo,semantic}. The shared view seems to be that employment of adversarial training and/or sophisticated loss such as the perceptual loss tends to deliver sharper and more realistic images, while their pixel-to-pixel differences (e.g., PSNR) from their ground truths tend not to be smaller (or sometimes even larger). In this study, however, we stick to 
the pixel-to-pixel difference due to the following reasons. First, which evaluation measure should be used depends on for which purpose we use these ``image restoration'' methods. For a photo-editing software, looking more photo-realistic will be more important. For the purpose of `pure' image restoration in which the goal is to predict intensities of missing pixels, it will be more important that each pixel has a value closer to its ground truth (see examples of inpainting images of numbers in Figure~\ref{fig_inpaint}). Second, popular quality measures, such as mean opinion score (MOS), need human raters, and their values are not easy to reproduce particularly when there are only small differences between images under comparison. 
%We also take into account that previous studies use PSNR for a main performance measure. 
Finally, we also note that our method does sometimes provide sharper images compared to existing methods (see examples of inpainting images with random pixel masks in Figure~\ref{fig_inpaint}).

% section 3
\section{Evolutionary Convolutional Autoencoders}

\subsection{Search Space of Network Architectures}

We consider convolutional autoencoders (CAEs) which are built only on standard building blocks of ConvNets, i.e., convolutional layers with optional downsampling and skip connections.
We further limit our attention to {\em symmetric} CAEs such that their first half (encoder part) is symmetric to the second half (decoder part). We add a final layer to obtain images of fixed channels (i.e., single-channel gray-scale or three-channel color images) on top of the decoder part, for which either one or three filters of $3\times 3$ size are used. Therefore, specification of the encoder part of a CAE solely determines its entire architecture.
The encoder part can have an arbitrary number of convolutional layers up to a specified maximum.
Each convolutional layer can have an arbitrary number and size of (single-size) filters, and is followed by ReLU \cite{nair_rectified_2010}.
Additionally, it can have an optional skip connection \cite{srivastava_2015, res, red}, which connects the layer to its mirrored counterpart in the decoder part.
To be specific, the output feature maps (obtained after ReLU) of the layer are passed to and are element-wise added to the output feature maps (obtained before ReLU) of the counterpart layer.
%Then, the maps are passed to the next layer following a  ReLU function.
We can use additional downsampling after each convolutional layer depending on tasks; whether to use downsampling is determined in advance, and thus is not selected by architectural search, as will be explained later.

\subsection{Representation of CAE Architectures}

Following \cite{suganuma}, we represent architectures of CAEs by directed acyclic graphs defined on a two-dimensional grid.
This graph is optimized by the evolutionary algorithm explained below, where the graph is called phenotype, and is encoded by a data structure called genotype \cite{Eiben}.

\begin{figure}[t]
%\vskip 0.2in
\begin{center}
\centerline{\includegraphics[scale=0.42]{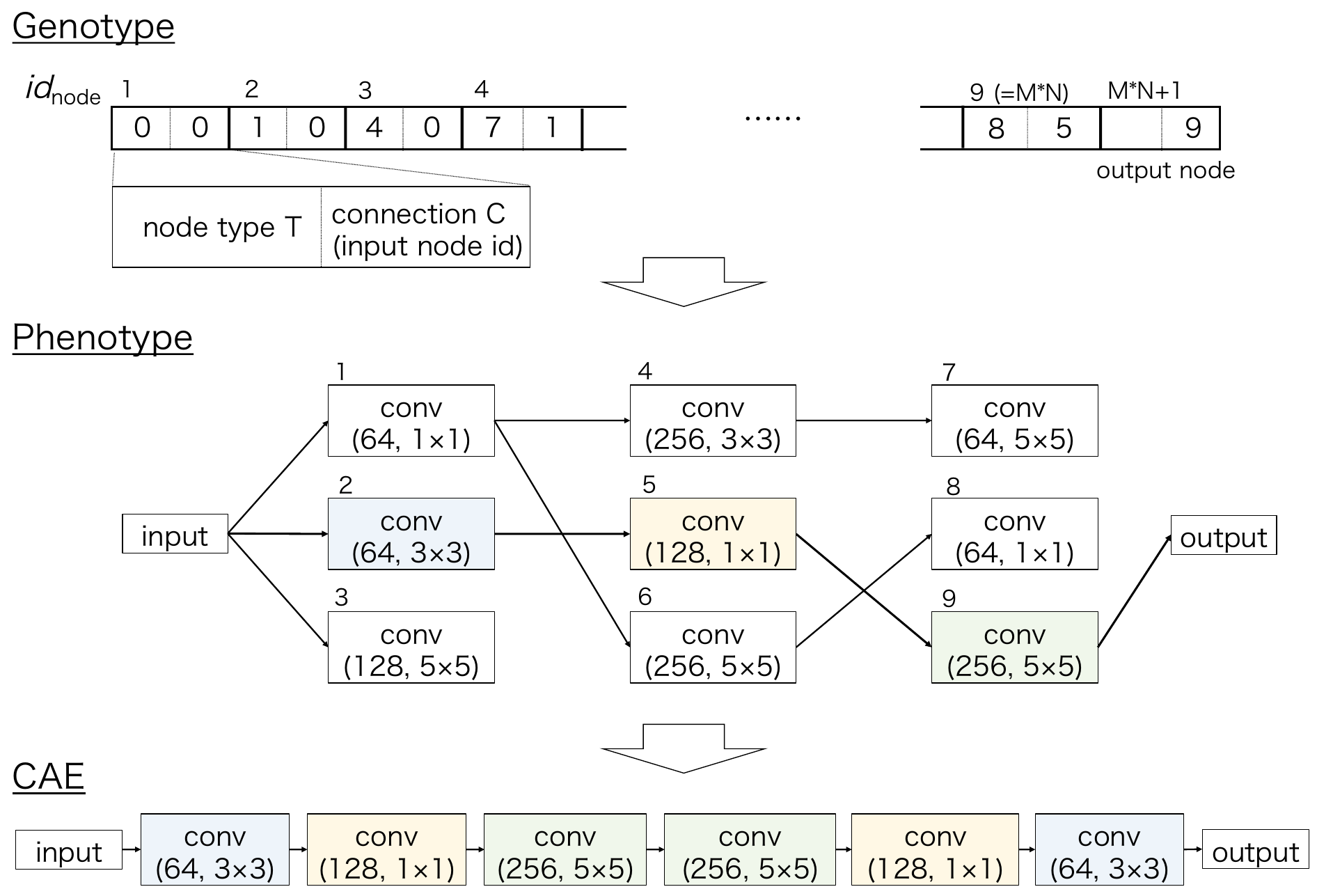}}
\caption{An example of a genotype and a phenotype.
A phenotype is a graph representation of a network architecture, and a genotype encodes a phenotype. They encode only the encoder part of a CAE, and its decoder part is automatically created so as to be symmetrical to the encoder part. In this example, the phenotype is defined on the grid of three rows and three columns. }
\label{genotype}
\end{center}
\vskip -0.2in
\end{figure}

\paragraph{Phenotype} 
A phenotype is a directed acyclic graph defined on a two-dimensional grid of $M$ rows and $N$ columns; see Fig.\ref{genotype}. Each node of the graph, which is identified by a unique $id_{\rm node}$ in the range $[1,MN]$ in a column-major order of the grid, represents a convolutional layer followed by a ReLU in a CAE. An edge connecting two nodes represents the connectivity of the two corresponding layers. The graph has two additional special nodes called  input and output nodes; the former represents the input layer of the CAE, and the latter represents the output of the encoder part, or equivalently the input of the decoder part of the CAE. 
%By traversing the graph from the output node to the input node, we have the architecture of the CAE, 
As the input of each node is connected to at most one node, there is a single unique path starting from the input node and ending at the output node. This unique path identifies the architecture of the CAE, as shown in the middle row of Figure~\ref{genotype}.
Note that nodes depicted in the neighboring two columns are not necessarily connected. Thus, the CAE can have  different number of layers depending on how their nodes are connected. Since the maximum number of layers (of the encoder part) of the CAE is $N$, the total number of layers is $2N+1$ including the output layer. In order to control how the number of layers will be chosen, we introduce a hyper-parameter called level-back $L$, such that nodes given in the $n$-th column are allowed to be connected from nodes given in the columns ranging from $n-L$ to $n-1$. If we use smaller $L$, then the resulting CAEs will tend to be deeper.

\paragraph{Genotype} A genotype encodes a phenotype, and is manipulated by the evolutionary algorithm. 
%It is a fixed length list of integers, which have information on a type and connections of nodes. 
The genotype encoding a phenotype with $M$ rows and $N$ columns has $MN+1$ genes, each of which represents attributes of a node  with two integers (i.e., type $T$ and connection $C$).
%the index ($id_{\rm{node}}$), 
% (This must be two integers, not three. I believe that each gene ``represents'' $T$ and $C$ alone. $id$ is just assigned to each gene, and gene does not represents $id$.)}
%{\color{red} three (two? only $T$ and $C$ as shown in Fig.; Besides "Node type \& Input node no." in the figure should be something like "$T$ \& $C$")} integers, the index ($id_{\rm{node}}$), type $T$, and connection $C$ of the node. 
%We give each gene the node index $id_{\rm{node}}$ in ascending order (e.g., $id_{\rm{node}} = 1, 2, 3,\ldots,MN$).
The type $T$ specifies the number $F$ and size $k$ of filters of the node, and whether the layer has skip connections or not, by an integer encoding their combination. The connection $C$ specifies the node by $id_{\rm{node}}$ that is connected to the input of this node. %The type and connections of the genotype are updated to maximize the model's quality on a validation dataset by evolutionary algorithms.
The last $(MN+1)$-st gene represents the output node, which stores only connection $C$ determining the node connected to the output node. An example of a genotype is given at the top row of Figure~\ref{genotype}, where $F\in\{64, 128, 256\}$ and $k\in\{1\times 1, 3\times 3, 5\times 5\}$.

\subsection{Evolutionary Strategy}
We use a simple form of the $(1+\lambda)$ evolutionary strategy \cite{miller_cartesian_2000} to perform search in the architecture space. 
%to search CAE architectures.
In this strategy, $\lambda$ children are generated from a single parent at each generation, and the best performing child compared to its parent becomes the new parent at the next generation. 
%we search the architecture space consisting of one parent and $\lambda$ children, which represent the genotype, i.e., CAE models.
The performance of each individual (i.e., a generated CAE), called {\em fitness}, is measured by 
%{\color{red} the $l_2$ distance} 
%The model's performance, e.g., 
 peak signal to noise ratio (PSNR)
between the restored and ground truth images evaluated on 
%the corrupted image and the output image, computed using a
the validation dataset.
%, is a measure of the individual's quality called fitness.
The genotype is updated to maximize the fitness as  generation proceeds.
%during the evolution process of the algorithm.

The details are given in Algorithm \ref{alg}. The algorithm starts with an initial parent, which is chosen to be a minimal CAE having a single convolution layer and a single deconvolution layer. 
%It is trained using the training set and its performance is evaluated on given fitness. 
%We then train the network represented by the parent followed by assigning the validation performance as the fitness.

At each generation, $\lambda$ children are generated by applying mutations to the parent (line 5). We use a point mutation as the genetic operator, where integer values of the type $T$ and connection $C$ of each gene are randomly changed with a mutation probability $r$. If a gene is decided to be changed, the mutation operator chooses a value at random for each $T$ and $C$ from  their predefined sets. 

The generated $\lambda$ children are individually trained using the training set. We train each child for $I$ iterations using the ADAM optimizer \cite{adam} with learning rate $lr$, and a mini-batch size of $b$ (line 6).
For the training loss, we use the mean squared error (MSE) between the restored images and their ground truths:
\begin{eqnarray}
L(\theta_{D}) = \frac{1}{|S|} \sum_{i=1}^{|S|} ||D(y_i;\theta_{D})-x_i||^2_{2},
\end{eqnarray}
where we denote the CAE and its weight parameters by $D$ and $\theta_{D}$, $S$ is the training set, $x_i$ is a ground truth image, and $y_i$ is a corrupted image.
After the training phase is completed, the performance of each child is evaluated on the validation set and is assigned to its fitness value (line 7). Finally, the best individual is selected from the set of parent and the children, and replaced the parent in the next generation (line $9-12$). This procedure is repeated for $G$ generations.

We can obtain a single unique path starting from the input node and ending at the output node using our representation.
The computed unique path represents the architecture of the CAE.
We call nodes on this path functioning nodes.
As some (in fact, most) of nodes in a phenotype are not functioning nodes and do not express the resulting CAE, the mutation has the possibility of affecting only non-functioning nodes, i.e., the CAE architecture does not change by the mutation.
%it is possible that a single application of the mutation operator does not change the CAE architecture.
In that case, we skip the evaluation of the CAE and apply the mutation operator repeatedly until the resulting CAE architecture does change. 
Moreover, if the fitness values of the children do not improve, then we modify a parent \cite{miller_cartesian_2000,miller_redundancy_2006}; in this case, we change only the non-functioning nodes so that the realized CAE (i.e., functioning nodes) will not change (line 14).

\begin{algorithm}[t]
   \caption{Evolutionary strategy for a CAE.}
   \label{alg}
\begin{algorithmic}[1]
   \STATE {\bfseries Input:} $G$ (number of generations), $r$ (mutation probability), $\lambda$ (children size), $S$ (Training set), $V$ (Validation set).
   \STATE {\bfseries Initialization:} (i) Generate a $parent$, (ii) train the model on the $S$, and (iii) assign the fitness $F_p$ using the set $V$.
   \WHILE{$generation$ $<$ $G$}
   \FOR{$i=1$ {\bfseries to} $\lambda$}
   \STATE $children_i$ $\leftarrow$ Mutation($parent$, $r$)
   \STATE $model_i$ $\leftarrow$ Train($children_i, S$)
   \STATE $fitness_i$ $\leftarrow$ Evaluate($model_i, V$)
   \ENDFOR
   \STATE $best$ $\leftarrow$ $\rm{argmax}_{i=1,2,\ldots,\lambda} \{ fitness_i \}$
   \IF{$f_{best} > F_{p}$}
   \STATE $parent \leftarrow children_{best}$
   \STATE $F_p \leftarrow f_{best}$
   \ELSE
   \STATE $parent$ $\leftarrow$ Modify($parent, r$)
   \ENDIF
   \STATE $generation = generation + 1$
   \ENDWHILE
   \STATE {\bfseries Output:} $parent$ (the best architecture of CAEs found by the evolutionary search).
   % \UNTIL{$noChange$ is $true$}
\end{algorithmic}
\end{algorithm}

% Section 4
\section{Experimental Results}
We conducted experiments to test the effectiveness of our approach. 
%To be specific, for each of the tasks and datasets that will be explained below, we use the proposed evolutionary algorithm to search for good CAE models and then fine-tune the best-performing model for additional xx epochs. 
%In what follows, we first compare the best CAE model with existing methods in terms of accuracy. We then analyze how the evolutionary algorithm explores CAE models. For this purpose we
We chose two tasks, image inpainting and denoising. %We first describe the experimental procedures for each task. 

%In this section, we examine our proposed methods, and compare the results with state-of-the-art methods for image inpainting and denoising tasks.

\subsection{Details of Experiments}
%Datasets and Details of Experiments}

\subsubsection{Inpainting}

We followed the procedures suggested in \cite{semantic} for experimental design. We used three benchmark datasets; the CelebFaces Attributes Dataset (CelebA) \cite{liu_deep_2015}, the Stanford Cars Dataset (Cars) \cite{cars}, and the Street View House Numbers (SVHN) \cite{svhn}. The CelebA contains $202,599$ images, from which we randomly selected $100,000$, 1,000, and 2,000 images for training, validation, and test, respectively. 
All images were cropped in order to properly contain the entire face, and resized to $64\times 64$ pixels.
For Cars and SVHN, we used the provided training and testing split.
The images of Cars were cropped according to the provided bounding boxes, and resized to $64\times 64$ pixels.
The images of SVHN were resized to $64\times 64$ pixels.

We generated images with missing regions of the following three types: a central square block mask ({\em Center}), random pixel masks such that 80\% of all the pixels were randomly masked ({\em Pixel}), and half image masks such that a randomly chosen vertical or horizontal half of the image was masked ({\em Half}). For the latter two, a mask was randomly generated for each training minibatch and for each test image. 

Considering the nature of this task, we consider CAEs endowed  with downsampling. To be specific, the same counts of downsampling and upsampling with stride $=2$ were employed such that the entire network has a symmetric hourglass shape. For simplicity, we used a skip connection and downsampling in an exclusive manner; in other words, every layer (in the encoder part) employed either a skip connection or downsampling.

\subsubsection{Denoising} 

We followed the experimental procedures described in \cite{red, mem}. We used gray-scale $300$ and $200$ images belonging to the BSD500 dataset \cite{BSD} to generate training and test images, respectively. For each image, we randomly extracted $64\times 64$ patches, to each of which Gaussian noise with different $\sigma=30, 50$ and $70$ are added. 
%In order to increase the number of samples, noise is generated and added to each training image multiple times. 
As utilized in the previous studies, we trained a single model for each different noise level.

For this task, we used CAE models without downsampling following the previous studies \cite{red, mem}. We zero-padded the input feature maps computed in each convolution layer not to change the size of input and output feature space of the layer. 

\subsection{Configurations of Architectural Search}

For the proposed evolutionary algorithm, we chose the mutation probability as $r=0.1$, the number of children as  $\lambda=4$, and the number of generations as $G=250$. For the phenotype, we used the graph with $M=3$, $N=20$ and level-back $L=5$. For the number $F$ and size $k$ of filters at each layer, we chose them from $\{64, 128, 256\}$ and $\{1\times 1, 3\times 3, 5\times 5\}$, respectively. During an evolution process, we trained each CAE for $I=20$k iterations with a mini-batch of size $b=16$.
We set the learning rate $lr$ of the ADAM optimizer to be $0.001$.
Following completion of the evolution process, we fine-tuned the best CAE using the training set of images for additional $500$k iterations, in which the learning rate is reduced by a factor of $10$ at the $200$k and $400$k iterations.
We then calculated its performance using the test set of images. We implemented our method using PyTorch \cite{paszke2017automatic}, and performed the experiments using four P100 GPUs. Execution of the evolutionary algorithm and the fine-tuning of the best model took about three days for the inpainting tasks and four days for the denoising tasks.

\begin{figure*}[t]
\begin{center}
\centerline{\includegraphics[scale=0.62]{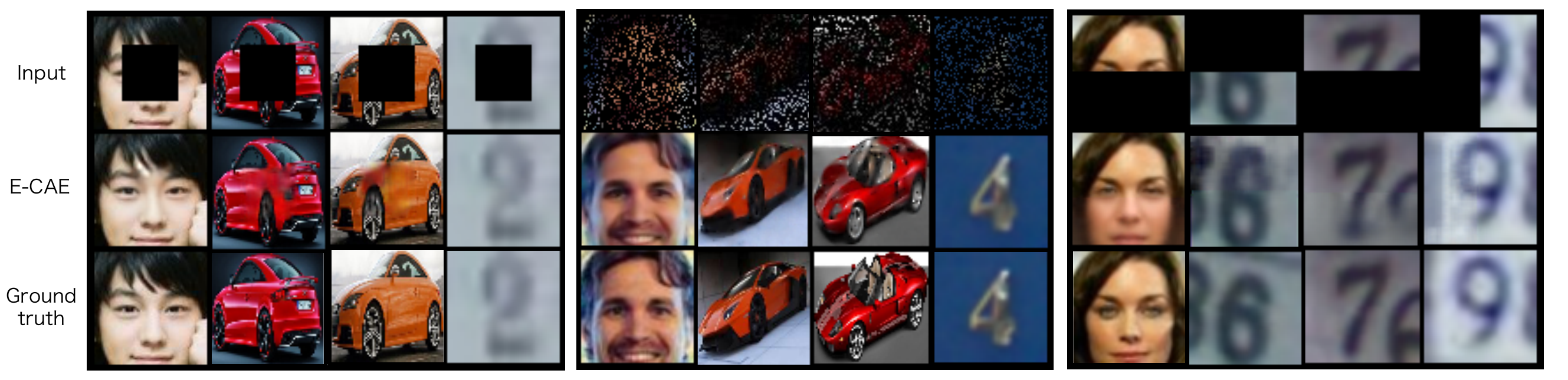}}
\caption{
Examples of inpainting results obtained by E-CAE (CAEs designed by the evolutionary algorithm).
}
\label{fig_inpaint}
\end{center}
\vskip -0.2in
\end{figure*}

\subsection{Comparison with Existing Methods} 

\subsubsection{Inpainting}

As mentioned above, we follow the experimental procedure employed in
\cite{semantic}. In the paper, the authors reported the performances of
their proposed method, Semantic Image Inpainting (SII), and Context
Autoencoder (CE) \cite{pathak2016context}. Borrowing the numbers
reported in the paper, it is straightforward to compare the performance of our approach against these two methods. However, we found that CE can provide considerably better results than those reported in \cite{semantic} in terms of both PSNR
and visual quality. Thus, we report here PSNR and SSIM values of CE that we obtained by running the authors' code\footnote{ https://github.com/pathak22/context-encoder}. In order to calculate SSIM values of SII, which were not reported in \cite{semantic}, we run the authors' code\footnote{https://github.com/moodoki/semantic\_image\_inpainting} for SII.

\begin{table}[t]
\caption{{\bf Inpainting results.} Comparison of Context Autoencoder (CE) \cite{pathak2016context}, Semantic Image Inpainting (SII) \cite{semantic}, and CAEs designed by our evolutionary algorithm (E-CAE) using three datasets and three masking patterns.}
%All the numbers are PSNR and SSIM. }
\label{results_inpainting}
%\vskip 0.15in
\begin{center}
\begin{small}
\begin{sc}
\begin{tabular}{ccccc} \hline
                       \multicolumn{5}{c}{PSNR}  \\ \hline
Dataset       & type&   CE & SII    & E-CAE \\ \hline
              & center & $28.5$ & $19.4$ & $\bf{29.9}$ \\ 
CelebA        & pixel  & $22.9$ & $22.8$ & $\bf{27.8}$ \\
              & half   & $19.9$ & $13.7$ & $\bf{21.1}$ \\ \hline
              & center & $19.6$ & $13.5$ & $\bf{20.9}$ \\ 
Cars          & pixel  & $15.6$ & $18.9$ & $\bf{19.5}$ \\
              & half   & $14.8$ & $11.1$ & $\bf{16.2}$ \\ \hline
              & center & $16.4$ & $19.0$ & $\bf{33.3}$ \\ 
SVHN          & pixel  & $30.5$ & $33.0$ & $\bf{40.4}$ \\
              & half   & $21.6$ & $14.6$ & $\bf{24.8}$ \\ \hline \multicolumn{5}{c}{}\\[-.5em] \hline
                       \multicolumn{5}{c}{SSIM}      \\ \hline
Dataset       & type & CE   & SII & E-CAE \\ \hline
              & center & $0.912$ & $0.907$ & $\bf{0.934}$ \\ 
CelebA        & pixel  & $0.730$ & $0.710$ & $\bf{0.887}$ \\
              & half   & $0.747$ & $0.582$ & $\bf{0.771}$ \\ \hline
              & center & $0.767$ & $0.721$ & $\bf{0.846}$ \\ 
Cars          & pixel  & $0.408$ & $0.412$ & $\bf{0.738}$ \\
              & half   & $0.576$ & $0.525$ & $\bf{0.610}$ \\ \hline
              & center & $0.791$ & $0.825$ & $\bf{0.953}$ \\ 
SVHN          & pixel  & $0.888$ & $0.786$ & $\bf{0.969}$ \\
              & half   & $0.756$ & $0.702$ & $\bf{0.848}$ \\ \hline
\end{tabular}
\end{sc}
\end{small}
\end{center}
\vskip -0.1in
\end{table}

Table \ref{results_inpainting} shows the PSNR and SSIM values obtained using 
three methods on three datasets and three masking patterns. Our method
(i.e., the CAE optimized by the evolutionary algorithm) is referred 
as E-CAE. We run the evolutionary algorithm three times, and report the
average accuracy values of the three optimized CAEs. As we can
see, our method outperforms the other two methods for each of the
dataset-mask combinations. It should also be noted that CE and SII
use mask patterns for inference; to be specific, their networks
estimate only pixel intensities of the missing regions specified by
the provided masks, and then they are merged with the unmasked regions
of clean pixels. Thus, the pixel intensities of unmasked
regions are identical to their ground truths. On the other
hand, our method does not use masks; it outputs a complete image such
that the missing regions are hopefully inpainted correctly. We then
calculate the PSNR of the output image against the ground truth
without identifying missing regions. This difference should favor CE
and SII, and nevertheless our method performs better.

\begin{figure}[t]
\begin{center}
\centerline{\includegraphics[scale=0.6]{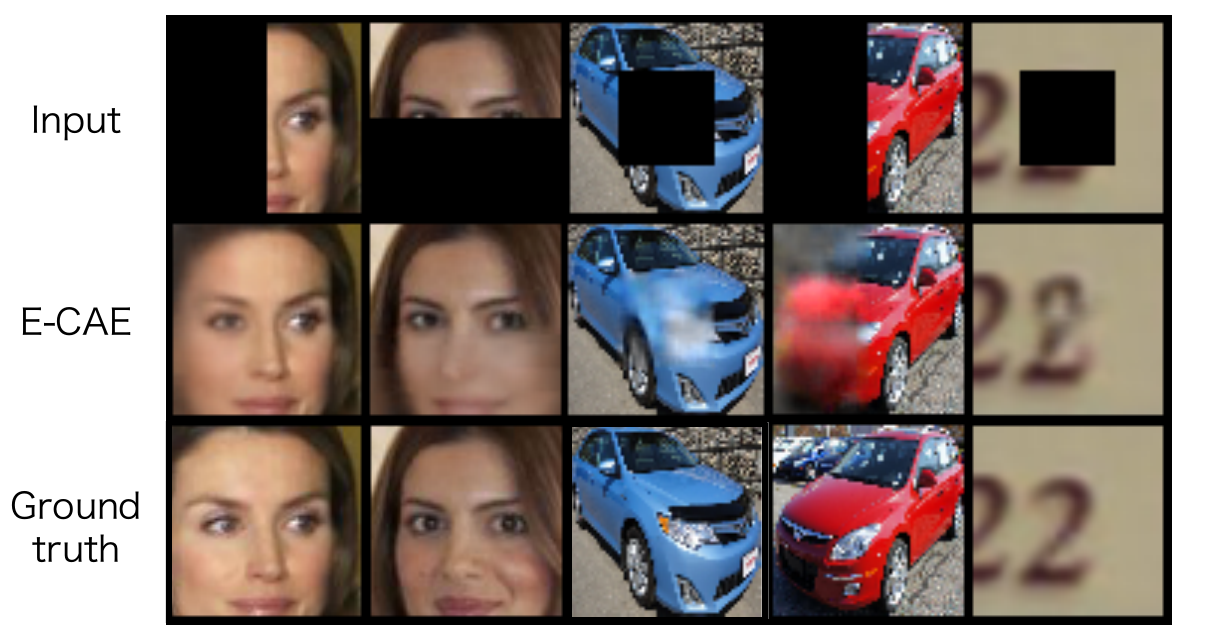}}
\caption{Examples of blurry reconstructions generated by E-CAE.}
\label{fig_inpaint_failure}
\end{center}
\vskip -0.2in
\end{figure}

Sample inpainted images obtained by E-CAE along with the masked inputs, and the ground truths are shown in Figure~\ref{fig_inpaint}. As we choose the same images as those used in \cite{semantic}, the readers can easily check differences in visual quality from CE and SII. It is observed overall that E-CAE performs stably; the output images do not have large errors for all types of masks. It performs particularly well for random pixel masks (the middle column of Figure~\ref{fig_inpaint}); the images are realistic and sharp. It is also observed that E-CAE tends to yield less sharp images for images with a filled region of missing pixels. However, E-CAE can infer their contents accurately, as shown in the examples of inpainting images of numbers (the rightmost column of Figure~\ref{fig_inpaint}); CE and SII provide either obscure images of numbers which are difficult to recognize, or sharp images of  wrong numbers; see Figure 18 and 21 of \cite{semantic_arXiv}.
Figure~\ref{fig_inpaint_failure} shows several examples of difficult cases for E-CAE.
We provide other examples of inpainting results in Figure~\ref{fig_inpaint_arXiv}.

\begin{table}[t]
\caption{\bf{Denoising results on BSD200.} \rm{Comparison of results of the BM3D \cite{dabov_bm3d_2009}, RED \cite{red}, MemNet \cite{mem}, and E-CAE. 
%All the numbers are average PSNR and SSIM. 
}}
\label{results_denoising}
%\vskip 0.15in
\begin{center}
\begin{small}
\begin{sc}
\begin{tabular}{ccccc} \hline
\multicolumn{5}{c}{PSNR}  \\ \hline
Noise $\sigma$ &  BM3D & RED & MemNet & E-CAE \\ \hline
$30$ & $27.31$ & $27.95$ & $28.04$ & $\bf{28.23}$ \\ 
$50$ & $25.06$ & $25.75$ & $25.86$ & $\bf{26.17}$ \\
$70$ & $23.82$ & $24.37$ & $24.53$ & $\bf{24.83}$ \\ 
\hline
\multicolumn{5}{c}{}\\[-.5em]
\hline
%              & $30$ & $28.49$ & $29.17$ & $\bf{29.22}$ & $28.75$ \\ 
%14 images& $50$ & $26.08$ & $26.81$ & $\bf{26.91}$ & $26.33$ \\
%              & $70$ & $24.65$ & $25.31$ & $\bf{25.43}$ & $24.68$ \\ \hline
\multicolumn{5}{c}{SSIM}  \\ \hline
Noise $\sigma$ &  BM3D    & RED     & MemNet  & E-CAE \\ \hline
$30$     & $0.7755$ & $0.8019$ & $\bf{0.8053}$ & $0.8047$ \\ 
$50$     & $0.6831$ & $0.7167$ & $0.7202$ & $\bf{0.7255}$ \\
$70$     & $0.6240$ & $0.6551$ & $0.6608$ & $\bf{0.6636}$ \\ \hline
%         & $30$     & $0.8204$ & $0.8423$ & $0.8444$ & $ $ \\ 
%14 images& $50$     & $0.7427$ & $0.7733$ & $0.7775$ & $ $ \\
%         & $70$     & $0.6882$ & $0.7206$ & $0.7260$ & $ $ \\ \hline
\end{tabular}
\end{sc}
\end{small}
\end{center}
\vskip -0.1in
\end{table}

\subsubsection{Denoising}

We compare our method with three 
%existing 
state-of-the-art methods; BM3D \cite{dabov_bm3d_2009}, RED \cite{red}, and MemNet \cite{mem}.
Table \ref{results_denoising} shows PSNR and SSIM values for three versions of the BSD200 test set with different noise levels $\sigma=30, 50$, and $70$, where the performance values of these three methods are obtained from \cite{mem}.
%since the previous studies strictly follows the same experimental procedure, and thus we can use those numbers 
Our method again achieves the best performance for all cases except a single case (MemNet for $\sigma=30$). It is worth noting that the networks of RED and MemNet have $30$ and $80$ layers, respectively, whereas our best CAE has only 15 layers (including the decoder part and the output layer), showing that our evolutionary method was able to find simpler architectures that can provide more accurate results.

\begin{figure}[t]
\begin{center}
\centerline{\includegraphics[scale=0.33]{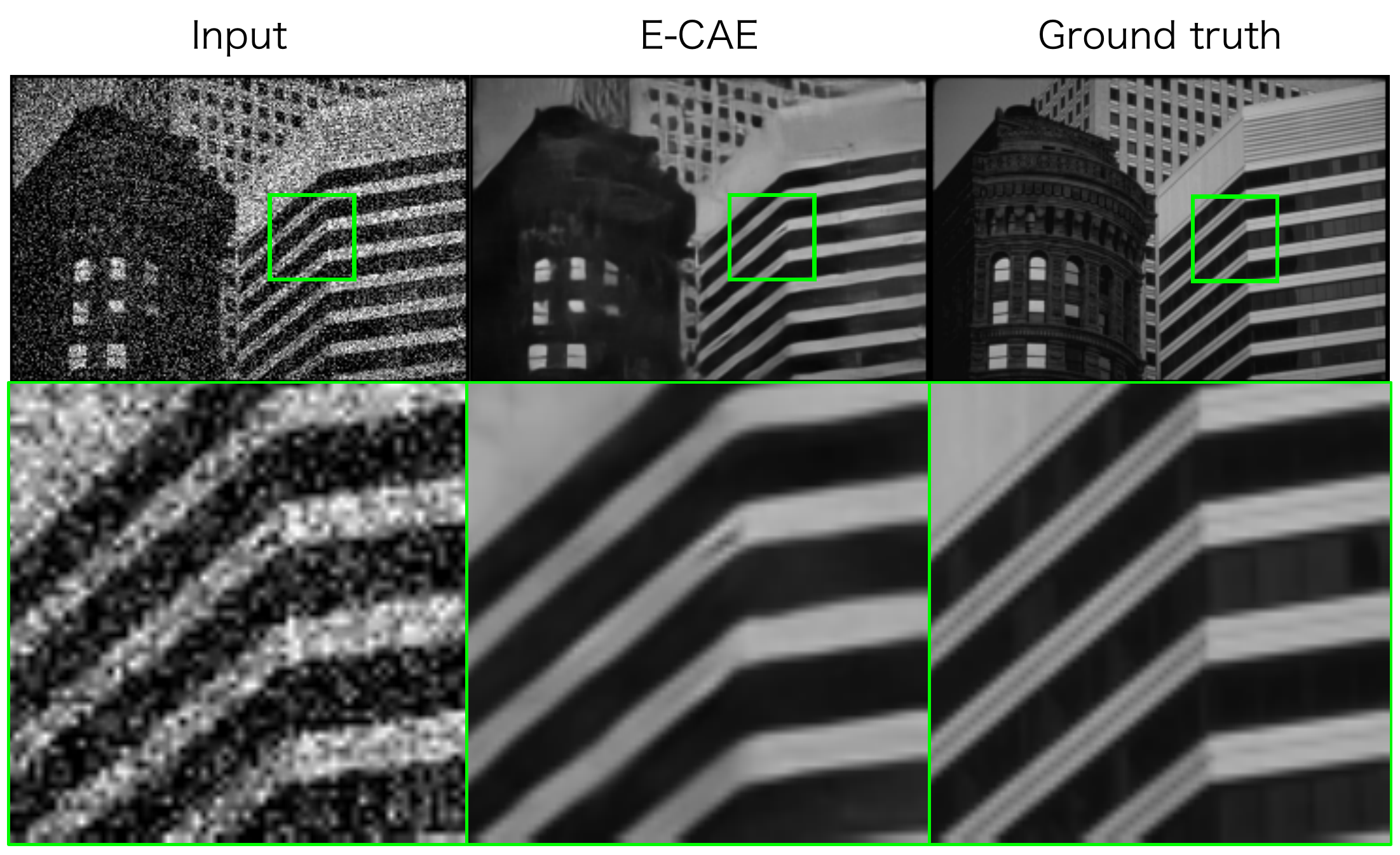}}
\caption{Examples of images reconstructed by E-CAE for the denoising task. The first column shows the input image with noise level $\sigma=50$. }
\label{fig_denoise}
\end{center}
\vskip -0.2in
\end{figure}

An example of an image recovered by our method is shown in Figure~\ref{fig_denoise}. As we can see, E-CAE correctly removes the noise, and produces an image as sharp as the ground truth.
We provide other examples of images reconstructed by E-CAE in Figure~\ref{fig_denoise_arXiv}.

\subsection{Analysis of Optimized Architectures}

\begin{figure*}[t]
\begin{center}
\smallskip
\centerline{\includegraphics[scale=0.6]{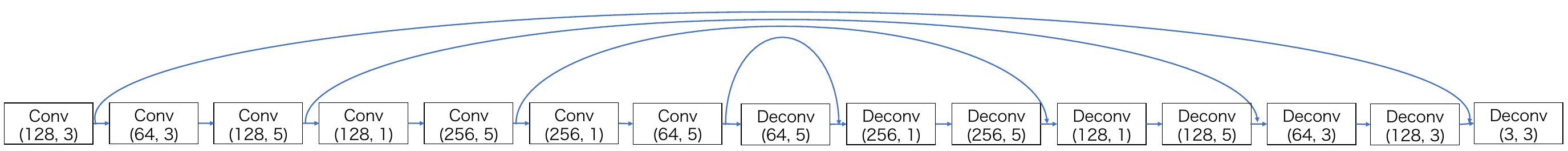}}
\medskip
\centerline{\includegraphics[scale=0.6]{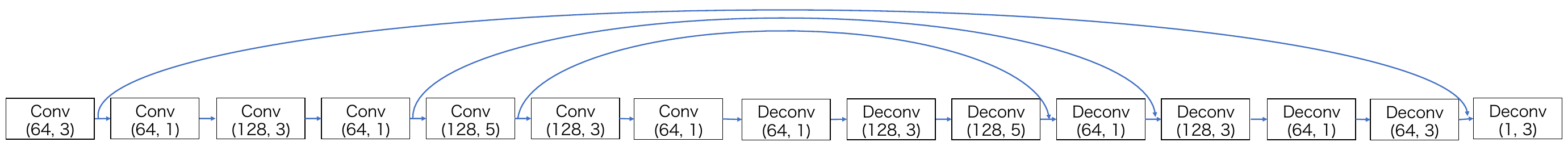}}
\caption{One of the best performing architectures given in Table \ref{top_five} for  inpainting (upper) and  denoising (lower) tasks. }
%Conv$(F,k)$ denotes that the layer has $F$ filters with the receptive filed of size $k\times k$. It is observed that although the overall structure appears to be rather standard, each layer has different parameters that are hard to choose manually. }
\label{architecture_inpaint}
\end{center}
\vskip -0.2in
\end{figure*}

Table \ref{top_five} shows the top five best performing architectures designed by our method for the image inpainting and denoising tasks, along with their performances measured on their test datasets. One of the architectures best performing for each task is shown in Figure~\ref{architecture_inpaint}. It is observed that although their overall structures do not look very unique, mostly due to the limited search space of CAEs, {\em the number and size of filters are quite different across layers, which are hard to manually determine}. Although it is difficult to give a general interpretation of why the parameters of each layer are chosen, we can make the following observations: i) regardless of the tasks, almost all networks have a skip connection at the first layer, implying that the input images contain essential information to yield accurate outputs; 
%contributes to the performance of the both tasks because the skip connection of the first layer can directly deliver the input feature to the output as shown in Fig. \ref{architecture_inpaint}.
ii) $1\times 1$ convolution seems to be important ingredients for both tasks; $1\times 1$ conv. layers dominate the denoising networks, and all the inpainting networks employ two $1\times 1$ conv. layers;
iii) when comparing the inpainting networks with the denoising networks, we observe the following differences: the largest filters of size $5\times 5$ tend to be employed by the former more often  than the latter (2.8 vs 0.8 layers in average), and $1\times 1$ filters tend to be employed by the former less often than the latter (2.0 vs. 3.2 layers in average).

\begin{table*}[t]
\caption{{\bf Best performing five architectures of E-CAE.} $C(F,k)$ indicates that the layer has $F$ filters of size $k\times k$ without a skip connection. $CS$ indicates that the layer has a skip connection. This table shows only the encoder part of CAEs. For the denoising, the average values of PSNR and SSIM of three noise levels are shown.}
\label{top_five}
%\vskip 0.15in
\medskip
\begin{small}
\begin{sc}
\hfil\begin{tabular}{p{140mm}|c|c} \hline
\bf{Architecture (Inpainting)} & PSNR & SSIM  \\ \hline
$CS(128,3)-C(64,3)-CS(128,5)-C(128,1)-CS(256,5)-C(256,1)-CS(64,5)$           & $29.91$ & $0.9344$ \\ \hline
$C(256,3)-CS(64,1)-C(128,3)-CS(256,5)-CS(64,1)-C(64,3)-CS(128,5)$            & $29.91$ & $0.9343$ \\ \hline
$CS(128,5)-CS(256,3)-C(64,1)-CS(128,3)-CS(64,5)-CS(64,1)-C(128,5)-C(256,5)$  & $29.89$ & $0.9334$ \\ \hline
$CS(128,3)-CS(64,3)-C(64,5)-CS(256,3)-C(128,3)-CS(128,5)-CS(64,1)-CS(64,1)$  & $29.88$ & $0.9346$ \\ \hline
$CS(64,1)-C(128,5)-CS(64,3)-C(64,1)-CS(256,5)-C(128,5)$                      & $29.63$ & $0.9308$ \\ \hline
\end{tabular}
\end{sc}

\medskip
\begin{sc}
\hfil\begin{tabular}{p{140mm}|c|c} \hline
%\vskip{-.5em}
\bf{Architecture (Denoising)} & PSNR & SSIM  \\ \hline
$CS(64,3)-C(64,1)-C(128,3)-CS(64,1)-CS(128,5)-C(128,3)-C(64,1)$            & $26.67$ & $0.7313$ \\ \hline
$CS(64,5)-CS(256,1)-C(256,1)-C(64,3)-CS(128,1)-C(64,3)-CS(128,1)-C(128,3)$ & $26.28$ & $0.7113$ \\ \hline
$CS(64,3)-C(64,1)-C(128,3)-CS(64,1)-CS(128,5)-C(128,3)-C(64,1)$            & $26.28$ & $0.7107$ \\ \hline
$CS(128,3)-CS(64,1)-C(64,3)-C(64,3)-CS(64,1)-C(64,3)$                      & $26.20$ & $0.7047$ \\ \hline
$CS(64,5)-CS(128,1)-CS(256,3)-CS(128,1)-CS(128,1)-C(64,1)-CS(64,3)$        & $26.18$ & $0.7037$ \\ \hline

\end{tabular}
\end{sc}
\end{small}
%\vskip -0.1in
\end{table*}

\subsection{Effects of Parameters of Evolutionary Search}
%Analysis of Effect of Meta-Parameters on the Performance}

The evolutionary algorithm has several parameters, two of which, i.e., the mutation probability ($r$) and the number of children ($\lambda$),  tend to have particularly large impact on the performance of the optimized E-CAEs.
%generated during the evolution process.
%Below, we discuss their relationship to the performance.
Using the center mask inpainting task on the CelebA dataset, we analyze their impact in detail in this subsection. 
%{\color{blue} (Which do we consider here performances of CAEs before or after their fine-tuning?)}

\begin{figure*}[!h]
\begin{center}
\centerline{\includegraphics[scale=1.25]{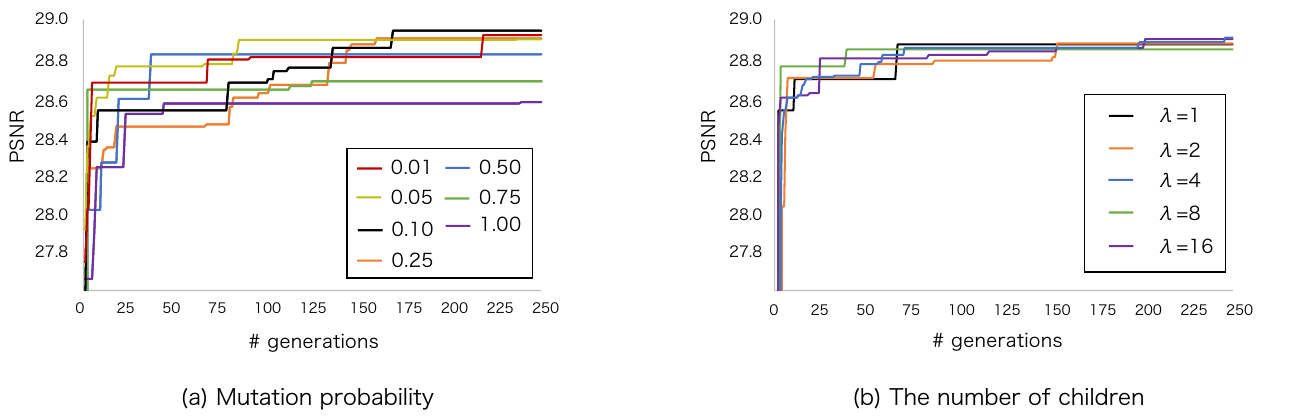}}
\caption{Improvement of PSNR of E-CAE by increasing number of generations obtained using the evolutionary algorithm for (a) different mutation probabilities, and (b) different number of children. The center mask inpainting task on the CelebA dataset is used. PSNR is calculated using the validation set. }
%using the validation set. Lower mutation rate shows better PSNR results on the validation set.}
\label{meta}
\end{center}
\vskip -0.2in
\end{figure*}

\paragraph{Effect of mutation probability} 
Employment of a larger mutation probability ($r$) will change the structures of CAEs more drastically at each generation, and  make the process of architecture search less stable.
On the other hand, a large mutation probability will contribute to reduce the possibility of being trapped in local optima.
Figure \ref{meta} (a) shows the relation between different mutation probabilities and the performances of CAEs obtained by using them; their performances are calculated on the validation set.
It is observed from the plots that smaller mutation probabilities tend to deliver lower accuracy at initial generations, but eventually provide higher accuracy after a sufficient number of generations are generated.
The best result was obtained for $r=0.1$.

\paragraph{Effect of number of children} 
Employment of a larger number ($\lambda$) of children will enable us to perform search in a wider subspace of the architecture space at each generation, but at the expense of larger computational cost per generation.
%If we use large number of children, then we can search the architecture space thoroughly, which helps to find better solutions.
Figure \ref{meta} (b) shows the relation between different $\lambda$ values ($\lambda=1, 2, 4, 8, $ and $16$) and the performances of the optimized CAEs.
The best performance is obtained for $\lambda=4$ using a sufficient number of generations, but there is not much difference in the final PSNR results obtained by different number of children.
Interestingly, even the evolution performed using $\lambda=1$, which uses the minimum computational cost per generation, yields a competitive result.
Specifically, it took $1.68$ days on one P100 GPU for training, and achieved PSNR $=29.80$ on the test set after fine-tuning of the model.

\section{Conclusion}

In this paper, we have first introduced an evolutionary algorithm that searches for \textit{good} architectures of convolutional autoencoders (CAEs) for image restoration tasks. We have then shown that the CAEs designed by our algorithm outperform the state-of-the-art networks for image inpainting and denoising, despite the fact that these networks are built on combination of complicated architectures with very deep layers, (multiple) hand-designed losses, and adversarial training; our CAEs consist only of standard convolutional layers with optional skip connections, and they are simply trained by the ADAM optimizer to minimize standard $\ell_2$ loss. Although our CAEs have simple architectures, their space is still very high-dimensional; CAEs can have an arbitrary number of layers, each of which has an arbitrary number and size of filters as well as whether to use a skip connection. Our evolutionary algorithm can find good architectures in this high-dimensional space. This implies that there is still much room for exploration of search spaces of  architectures of classical convolutional networks, which may apply to other tasks such as single image colorization \cite{zhang2016colorful}, depth estimation \cite{eigen2014depth,xu2017multi}, and optical flow estimation \cite{flownet}.


\bibliography{example_paper}

\begin{thebibliography}{54}
\providecommand{\natexlab}[1]{#1}
\providecommand{\url}[1]{\texttt{#1}}
\expandafter\ifx\csname urlstyle\endcsname\relax
  \providecommand{\doi}[1]{doi: #1}\else
  \providecommand{\doi}{doi: \begingroup \urlstyle{rm}\Url}\fi

\bibitem[Aharon et~al.(2006)Aharon, Elad, and Bruckstein]{ksvd}
Aharon, M., Elad, M., and Bruckstein, A.
\newblock k-{SVD}: {An} algorithm for designing overcomplete dictionaries for
  sparse representation.
\newblock \emph{IEEE Transactions on Signal Processing}, 54\penalty0
  (11):\penalty0 4311--4322, 2006.

\bibitem[Arora et~al.(2017)Arora, Ge, Liang, Ma, and
  Zhang]{Arora_ICML17_Generalization}
Arora, S., Ge, R., Liang, Y., Ma, T., and Zhang, Y.
\newblock Generalization and equilibrium in generative adversarial nets (gans).
\newblock In \emph{ICML}, pp.\  224--232, 2017.

\bibitem[Baker et~al.(2017)Baker, Gupta, Naik, and
  Raskar]{baker_designing_2016}
Baker, B., Gupta, O., Naik, N., and Raskar, R.
\newblock Designing neural network architectures using reinforcement learning.
\newblock In \emph{ICLR}, 2017.

\bibitem[Brock et~al.(2018)Brock, Lim, Ritchie, and Weston]{smash}
Brock, A., Lim, T., Ritchie, J.~M., and Weston, N.
\newblock Smash: one-shot model architecture search through hypernetworks.
\newblock In \emph{ICLR}, 2018.

\bibitem[Dabov et~al.(2009)Dabov, Foi, Katkovnik, and
  Egiazarian]{dabov_bm3d_2009}
Dabov, K., Foi, A., Katkovnik, V., and Egiazarian, K.
\newblock {BM}3d image denoising with shape-adaptive principal component
  analysis.
\newblock In \emph{SPARS}, 2009.

\bibitem[Dong et~al.(2014)Dong, Loy, He, and Tang]{dong2014learning}
Dong, C., Loy, C.~C., He, K., and Tang, X.
\newblock Learning a deep convolutional network for image super-resolution.
\newblock In \emph{ECCV}, pp.\  184--199, 2014.

\bibitem[Eiben \& Smith(2003)Eiben and Smith]{Eiben}
Eiben, A.~E. and Smith, J.~E.
\newblock \emph{Introduction to Evolutionary Computing}, volume~53.
\newblock SpringerVerlag, 2003.

\bibitem[Eigen et~al.(2014)Eigen, Puhrsch, and Fergus]{eigen2014depth}
Eigen, D., Puhrsch, C., and Fergus, R.
\newblock Depth map prediction from a single image using a multi-scale deep
  network.
\newblock In \emph{NIPS}, pp.\  2366--2374, 2014.

\bibitem[Fattal(2007)]{fattal}
Fattal, R.
\newblock Image upsampling via imposed edge statistics.
\newblock In \emph{SIGGRAPH}, 2007.

\bibitem[Goodfellow et~al.(2014)Goodfellow, Pouget-Abadie, Mirza, Xu,
  Warde-Farley, Ozair, Courville, and Bengio]{gan}
Goodfellow, I., Pouget-Abadie, J., Mirza, M., Xu, B., Warde-Farley, D., Ozair,
  S., Courville, A., and Bengio, Y.
\newblock Generative adversarial nets.
\newblock In \emph{NIPS}, pp.\  2672--2680, 2014.

\bibitem[He et~al.(2016)He, Zhang, Ren, and Sun]{res}
He, K., Zhang, X., Ren, S., and Sun, J.
\newblock Deep residual learning for image recognition.
\newblock In \emph{CVPR}, pp.\  770--778, 2016.

\bibitem[Iizuka et~al.(2017)Iizuka, Simo-Serra, and
  Ishikawa]{IizukaSIGGRAPH2017}
Iizuka, S., Simo-Serra, E., and Ishikawa, H.
\newblock Globally and locally consistent image completion.
\newblock In \emph{SIGGRAPH}, pp.\  107:1--107:14, 2017.

\bibitem[Ilg et~al.(2017)Ilg, Mayer, Saikia, Keuper, Dosovitskiy, and
  Brox]{flownet}
Ilg, E., Mayer, N., Saikia, T., Keuper, M., Dosovitskiy, A., and Brox, T.
\newblock Flownet 2.0: Evolution of optical flow estimation with deep networks.
\newblock In \emph{CVPR}, 2017.

\bibitem[Johnson et~al.(2016)Johnson, Alahi, and
  Fei-Fei]{Johnson2016Perceptual}
Johnson, J., Alahi, A., and Fei-Fei, L.
\newblock Perceptual losses for real-time style transfer and super-resolution.
\newblock In \emph{ECCV}, pp.\  694--711, 2016.

\bibitem[Kingma \& Ba(2015)Kingma and Ba]{adam}
Kingma, D.~P. and Ba, J.~L.
\newblock Adam: {A} method for stochastic optimization.
\newblock In \emph{ICLR}, 2015.

\bibitem[Krause et~al.(2013)Krause, Stark, Deng, and Fei-Fei]{cars}
Krause, J., Stark, M., Deng, J., and Fei-Fei, L.
\newblock 3d object representations for fine-grained categorization.
\newblock In \emph{ICCV Workshops (ICCVW)}, pp.\  554--561, 2013.

\bibitem[Krizhevsky et~al.(2012)Krizhevsky, Sutskever, and
  Hinton]{krizhevsky_imagenet_2012}
Krizhevsky, A., Sutskever, I., and Hinton, G.~E.
\newblock Imagenet classification with deep convolutional neural networks.
\newblock In \emph{NIPS}, pp.\  1097--1105, 2012.

\bibitem[Kulkarni et~al.(2016)Kulkarni, Lohit, Turaga, Kerviche, and
  Ashok]{kulkarni_reconnet_2016}
Kulkarni, K., Lohit, S., Turaga, P., Kerviche, R., and Ashok, A.
\newblock Reconnet: {Non}-iterative reconstruction of images from compressively
  sensed measurements.
\newblock In \emph{CVPR}, pp.\  449--458, 2016.

\bibitem[LeCun et~al.(1998)LeCun, Bottou, Bengio, and
  Haffner]{lecun_gradient_1998}
LeCun, Y., Bottou, L., Bengio, Y., and Haffner, P.
\newblock Gradient-based learning applied to document recognition.
\newblock \emph{Proceedings of the IEEE}, 86\penalty0 (11):\penalty0
  2278--2324, 1998.

\bibitem[Ledig et~al.(2017)Ledig, Theis, Husz{\'a}r, Caballero, Cunningham,
  Acosta, Aitken, Tejani, Totz, Wang, et~al.]{ledig2016photo}
Ledig, C., Theis, L., Husz{\'a}r, F., Caballero, J., Cunningham, A., Acosta,
  A., Aitken, A., Tejani, A., Totz, J., Wang, Z., et~al.
\newblock Photo-realistic single image super-resolution using a generative
  adversarial network.
\newblock In \emph{CVPR}, pp.\  4681--4690, 2017.

\bibitem[Liu et~al.(2017)Liu, Zoph, Shlens, Hua, Li, Fei-Fei, Yuille, Huang,
  and Murphy]{liu2017progressive}
Liu, C., Zoph, B., Shlens, J., Hua, W., Li, L., Fei-Fei, L., Yuille, A., Huang,
  J., and Murphy, K.
\newblock Progressive neural architecture search.
\newblock \emph{arXiv:1712.00559}, 2017.

\bibitem[Liu et~al.(2018)Liu, Simonyan, Vinyals, Fernando, and
  Kavukcuoglu]{hierarchical}
Liu, H., Simonyan, K., Vinyals, O., Fernando, C., and Kavukcuoglu, K.
\newblock Hierarchical representations for efficient architecture search.
\newblock In \emph{ICLR}, 2018.

\bibitem[Liu et~al.(2015)Liu, Luo, Wang, and Tang]{liu_deep_2015}
Liu, Z., Luo, P., Wang, X., and Tang, X.
\newblock Deep learning face attributes in the wild.
\newblock In \emph{ICCV}, pp.\  3730--3738, 2015.

\bibitem[Lucic et~al.(2017)Lucic, Kurach, Michalski, Gelly, and
  Bousquet]{arXiv1711.10337}
Lucic, M., Kurach, K., Michalski, M., Gelly, S., and Bousquet, O.
\newblock Are gans created equal? a large-scale study.
\newblock \emph{arXiv:1711.10337}, 2017.

\bibitem[Mao et~al.(2016)Mao, Shen, and Yang]{red}
Mao, X., Shen, C., and Yang, Y.
\newblock Image restoration using very deep convolutional encoder-decoder
  networks with symmetric skip connections.
\newblock In \emph{NIPS}, pp.\  2802--2810, 2016.

\bibitem[Martin et~al.(2001)Martin, Fowlkes, Tal, and Malik]{BSD}
Martin, D., Fowlkes, C., Tal, D., and Malik, J.
\newblock A database of human segmented natural images and its application to
  evaluating segmentation algorithms and measuring ecological statistics.
\newblock In \emph{ICCV}, pp.\  416--423, 2001.

\bibitem[Miikkulainen et~al.(2017)Miikkulainen, Liang, Meyerson, Rawal, Fink,
  Francon, Raju, Shahrzad, Navruzyan, Duffy, and
  Hodjat]{miikkulainen_evolving_2017}
Miikkulainen, R., Liang, J.~Z., Meyerson, E., Rawal, A., Fink, D., Francon, O.,
  Raju, B., Shahrzad, H., Navruzyan, A., Duffy, N., and Hodjat, B.
\newblock Evolving deep neural networks.
\newblock In \emph{GECCO}, 2017.

\bibitem[Miller \& Smith(2006)Miller and Smith]{miller_redundancy_2006}
Miller, J.~F. and Smith, S.~L.
\newblock Redundancy and computational efficiency in cartesian genetic
  programming.
\newblock \emph{IEEE Transactions on Evolutionary Computation}, 10\penalty0
  (2):\penalty0 167--174, 2006.

\bibitem[Miller \& Thomson(2000)Miller and Thomson]{miller_cartesian_2000}
Miller, J.~F. and Thomson, P.
\newblock Cartesian genetic programming.
\newblock In \emph{EuroGP}, pp.\  121--132, 2000.

\bibitem[Mousavi \& Baraniuk(2017)Mousavi and Baraniuk]{mousavi_learning_2017}
Mousavi, A. and Baraniuk, R.~G.
\newblock Learning to invert: {Signal} recovery via deep convolutional
  networks.
\newblock In \emph{ICASSP}, pp.\  2272--2276, 2017.

\bibitem[Nair \& Hinton(2010)Nair and Hinton]{nair_rectified_2010}
Nair, V. and Hinton, G.~E.
\newblock Rectified linear units improve restricted boltzmann machines.
\newblock In \emph{ICML}, pp.\  807--814, 2010.

\bibitem[Netzer et~al.(2011)Netzer, Wang, Coates, Bissacco, Wu, and Ng]{svhn}
Netzer, Y., Wang, T., Coates, A., Bissacco, A., Wu, B., and Ng, A.~Y.
\newblock Reading digits in natural images with unsupervised feature learning.
\newblock In \emph{NIPS workshop on deep learning and unsupervised feature
  learning}, 2011.

\bibitem[Paszke et~al.(2017)Paszke, Gross, Chintala, Chanan, Yang, DeVito, Lin,
  Desmaison, Antiga, and Lerer]{paszke2017automatic}
Paszke, A., Gross, S., Chintala, S., Chanan, G., Yang, E., DeVito, Z., Lin, Z.,
  Desmaison, A., Antiga, L., and Lerer, A.
\newblock Automatic differentiation in pytorch.
\newblock In \emph{NIPS Workshop}, 2017.

\bibitem[Pathak et~al.(2016)Pathak, Krahenbuhl, Donahue, Darrell, and
  Efros]{pathak2016context}
Pathak, D., Krahenbuhl, P., Donahue, J., Darrell, T., and Efros, A.~A.
\newblock Context encoders: Feature learning by inpainting.
\newblock In \emph{CVPR}, pp.\  2536--2544, 2016.

\bibitem[Perrone \& Favaro(2014)Perrone and Favaro]{perrone_total_2014}
Perrone, D. and Favaro, P.
\newblock Total variation blind deconvolution: {The} devil is in the details.
\newblock In \emph{CVPR}, pp.\  2909--2916, 2014.

\bibitem[Real et~al.(2017)Real, Moore, Selle, Saxena, Suematsu, Le, and
  Kurakin]{real_large_2017}
Real, E., Moore, S., Selle, A., Saxena, S., Suematsu, Y.~L., Le, Q.~V., and
  Kurakin, A.
\newblock Large-scale evolution of image classifiers.
\newblock In \emph{ICML}, pp.\  2902--2911, 2017.

\bibitem[Schaffer et~al.(1992)Schaffer, Whitley, and
  Eshelman]{schaffer_combinations_1992}
Schaffer, J.~D., Whitley, D., and Eshelman, L.~J.
\newblock Combinations of genetic algorithms and neural networks: {A} survey of
  the state of the art.
\newblock In \emph{COGANN}, pp.\  1--37, 1992.

\bibitem[Srivastava et~al.(2015)Srivastava, Greff, and
  Schmidhuber]{srivastava_2015}
Srivastava, R.~K., Greff, K., and Schmidhuber, J.
\newblock Training very deep networks.
\newblock In \emph{NIPS}, pp.\  2377--2385, 2015.

\bibitem[Stanley \& Miikkulainen(2002)Stanley and
  Miikkulainen]{stanley_evolving_2002}
Stanley, K.~O. and Miikkulainen, R.
\newblock Evolving neural networks through augmenting topologies.
\newblock \emph{Evolutionary Computation}, 10\penalty0 (2):\penalty0 99--127,
  2002.

\bibitem[Suganuma et~al.(2017)Suganuma, Shirakawa, and Nagao]{suganuma}
Suganuma, M., Shirakawa, S., and Nagao, T.
\newblock A genetic programming approach to designing convolutional neural
  network architectures.
\newblock In \emph{GECCO}, pp.\  497--504, 2017.

\bibitem[Tai et~al.(2017)Tai, Yang, Liu, and Xu]{mem}
Tai, Y., Yang, J., Liu, X., and Xu, C.
\newblock Memnet: {A} persistent memory network for image restoration.
\newblock In \emph{CVPR}, pp.\  4539--4547, 2017.

\bibitem[Xie et~al.(2012)Xie, Xu, and Chen]{xie_image_2012}
Xie, J., Xu, L., and Chen, E.
\newblock Image denoising and inpainting with deep neural networks.
\newblock In \emph{NIPS}, pp.\  341--349, 2012.

\bibitem[Xie \& Yuille(2017)Xie and Yuille]{xie_genetic_2017}
Xie, L. and Yuille, A.~L.
\newblock Genetic {CNN}.
\newblock In \emph{ICCV}, pp.\  1379--1388, 2017.

\bibitem[Xu et~al.(2017)Xu, Ricci, Ouyang, Wang, and Sebe]{xu2017multi}
Xu, D., Ricci, E., Ouyang, W., Wang, X., and Sebe, N.
\newblock Multi-scale continuous crfs as sequential deep networks for monocular
  depth estimation.
\newblock In \emph{CVPR}, pp.\  5354--5362, 2017.

\bibitem[Xu et~al.(2014)Xu, Ren, Liu, and Jia]{xu2014deep}
Xu, L., Ren, J.~SJ., Liu, C., and Jia, J.
\newblock Deep convolutional neural network for image deconvolution.
\newblock In \emph{NIPS}, pp.\  1790--1798, 2014.

\bibitem[Yang et~al.(2017)Yang, Lu, Lin, Shechtman, Wang, and Li]{yang2017high}
Yang, C., Lu, X., Lin, Z., Shechtman, E., Wang, O., and Li, H.
\newblock High-resolution image inpainting using multi-scale neural patch
  synthesis.
\newblock In \emph{CVPR}, pp.\  6721--6729, 2017.

\bibitem[Yang et~al.(2010)Yang, Wright, Huang, and Ma]{yang_image_2010}
Yang, J., Wright, J., Huang, T.~S., and Ma, Y.
\newblock Image super-resolution via sparse representation.
\newblock \emph{IEEE transactions on Image Processing}, 19\penalty0
  (11):\penalty0 2861--2873, 2010.

\bibitem[Yeh et~al.(2017{\natexlab{a}})Yeh, Chen, Lim, Schwing,
  Hasegawa-Johnson, and Do]{semantic}
Yeh, R.~A., Chen, C., Lim, T.~Y., Schwing, A.~G., Hasegawa-Johnson, M., and Do,
  M.~N.
\newblock Semantic image inpainting with deep generative models.
\newblock In \emph{CVPR}, pp.\  5485--5493, 2017{\natexlab{a}}.

\bibitem[Yeh et~al.(2017{\natexlab{b}})Yeh, Chen, Lim, Schwing,
  Hasegawa-Johnson, and Do]{semantic_arXiv}
Yeh, R.~A., Chen, C., Lim, T.~Y., Schwing, A.~G., Hasegawa-Johnson, M., and Do,
  M.~N.
\newblock Semantic image inpainting with deep generative models.
\newblock \emph{arXiv:1607.07539v3}, 2017{\natexlab{b}}.

\bibitem[Zhang. et~al.(2017)Zhang., Zuo, Chen, Meng, and
  Zhang]{zhang_beyond_2017}
Zhang., K., Zuo, W., Chen, Y., Meng, D., and Zhang, L.
\newblock Beyond a gaussian denoiser: {Residual} learning of deep cnn for image
  denoising.
\newblock \emph{IEEE Transactions on Image Processing}, 26\penalty0
  (7):\penalty0 3142--3155, 2017.

\bibitem[Zhang et~al.(2016)Zhang, Isola, and Efros]{zhang2016colorful}
Zhang, R., Isola, P., and Efros, A.~A.
\newblock Colorful image colorization.
\newblock In \emph{ECCV}, pp.\  649--666, 2016.

\bibitem[Zhong et~al.(2017)Zhong, Yan, and Liu]{zhong_practical_2017}
Zhong, Z., Yan, J., and Liu, C.
\newblock Practical network blocks design with {Q}-{Learning}.
\newblock In \emph{arXiv: 1708.05552}, 2017.

\bibitem[Zoph \& Le(2017)Zoph and Le]{zoph_neural_2016}
Zoph, B. and Le, Q.~V.
\newblock Neural architecture search with reinforcement learning.
\newblock In \emph{ICLR}, 2017.

\bibitem[Zoph et~al.(2017)Zoph, Vasudevan, Shlens, and Le]{zoph2017learning}
Zoph, B., Vasudevan, V., Shlens, J., and Le, Q.~V.
\newblock Learning transferable architectures for scalable image recognition.
\newblock \emph{arXiv:1707.07012}, 2017.

\end{thebibliography}
\bibliographystyle{icml2018}

\appendix
\section{Supplementary Material}
\begin{figure*}[t]
\begin{center}
\centerline{\includegraphics[scale=1.8]{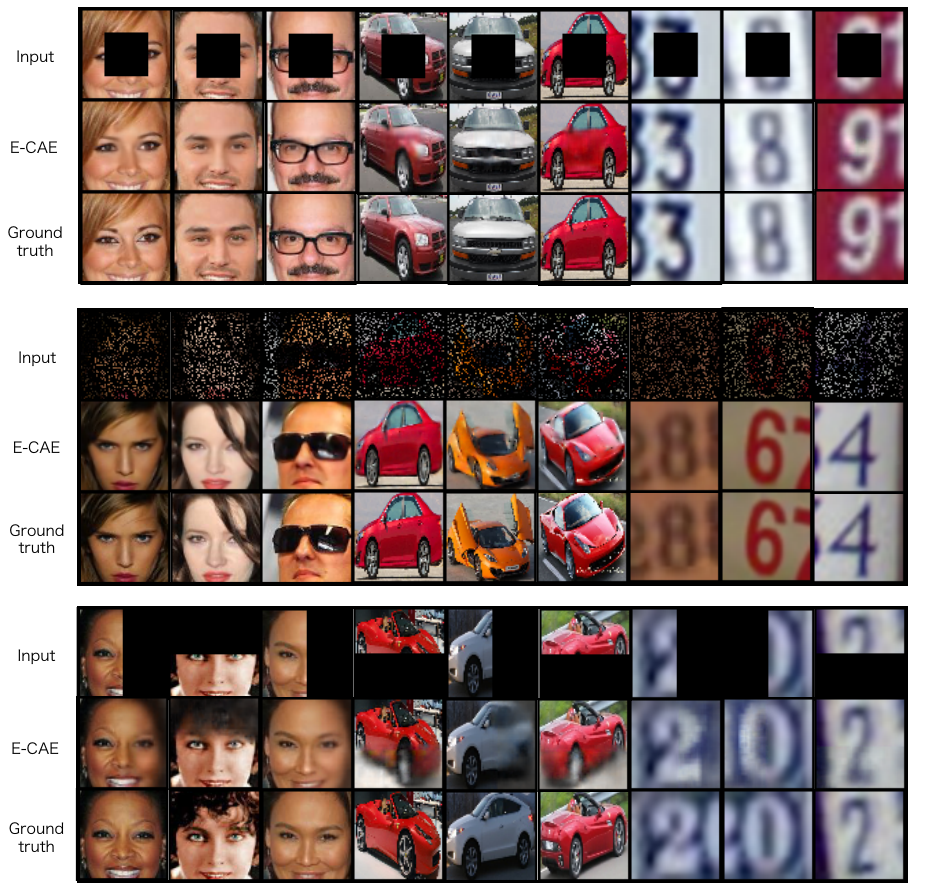}}
\caption{Examples of inpainting results obtained by E-CAE for distortion with the center mask (top row), pixel mask (middle row), and half mask (bottom row).}
\label{fig_inpaint_arXiv}
\end{center}
\vskip -0.2in
\end{figure*}

Figure \ref{fig_inpaint_arXiv} shows several inpainted images obtained by E-CAE along with the masked inputs and the ground truths.
The top row, the middle row, and the bottom row provide the results obtained for image distortion with center mask, the random pixel mask, and the half mask tasks, respectively.

\begin{figure*}[t]
\begin{center}
\centerline{\includegraphics[scale=0.6]{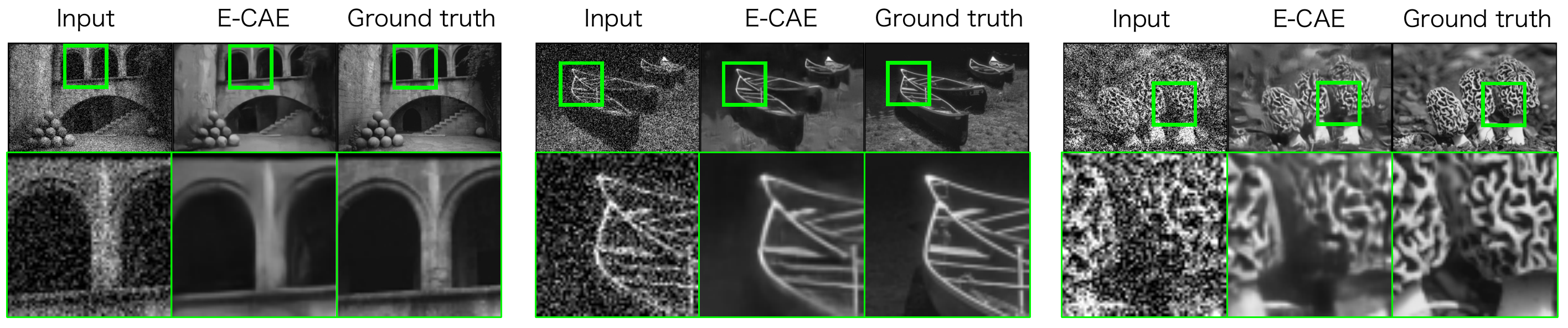}}
\caption{Examples of images reconstructed by E-CAE for the denoising task with noise level $\sigma=30$ (left column), $\sigma=50$ (middle column), and $\sigma=70$ (right column). }
\label{fig_denoise_arXiv}
\end{center}
\vskip -0.2in
\end{figure*}

For the denoising task, examples of images recovered by E-CAE are shown in Figure~\ref{fig_denoise_arXiv}.
The left column, the middle column, and the right column show the results for noise level $\sigma=30$, $\sigma=50$, and $\sigma=70$, respectively.

Our code is available at https://github.com/sg-nm/Evolutionary-Autoencoders.

%
%\textbf{\emph{Do not put content after the references.}}
%%
%Put anything that you might normally include after the references in a separate
%supplementary file.
%
%We recommend that you build supplementary material in a separate document.
%If you must create one PDF and cut it up, please be careful to use a tool that
%doesn't alter the margins, and that doesn't aggressively rewrite the PDF file.
%pdftk usually works fine. 
%
%\textbf{Please do not use Apple's preview to cut off supplementary material.} In
%previous years it has altered margins, and created headaches at the camera-ready
%stage. 
%%%%%%%%%%%%%%%%%%%%%%%%%%%%%%%%%%%%%%%%%%%%%%%%%%%%%%%%%%%%%%%%%%%%%%%%%%%%%%%
%%%%%%%%%%%%%%%%%%%%%%%%%%%%%%%%%%%%%%%%%%%%%%%%%%%%%%%%%%%%%%%%%%%%%%%%%%%%%%%

\end{document}